\documentclass[sigplan,authorversion,screen]{acmart}
\AtBeginDocument{%
  \providecommand\BibTeX{{%
    \normalfont B\kern-0.5em{\scshape i\kern-0.25em b}\kern-0.8em\TeX}}}

\setcopyright{none}

\renewcommand\footnotetextcopyrightpermission[1]{} 
\acmConference[May]{}{25}{2023}
\acmPrice{15.00}
\acmISBN{978-1-4503-XXXX-X/18/06}

\begin{document}

\title{Machine Learning to detect cyber-attacks and discriminating the types of power system disturbances}

\author{Diane Tuyizere}
\email{dtuyizer@andrew.cmu.edu}
\affiliation{%
  \institution{Carnegie Mellon University Africa}
  \city{Kigali}
  \country{Rwanda}
}
\author{Remy Ihabwikuzo}
\email{rihabwik@andrew.cmu.edu}
\affiliation{%
  \institution{Carnegie Mellon University Africa}
  \city{Kigali}
  \country{Rwanda}
}

\renewcommand{\shortauthors}{Diane et al.}

\begin{abstract}
  This research proposes a machine learning-based attack detection model for power systems, specifically targeting smart grids. By utilizing data and logs collected from Phasor Measuring Devices (PMUs), the model aims to learn system behaviors and effectively identify potential security boundaries. The proposed approach involves crucial stages including dataset pre-processing, feature selection, model creation, and evaluation. To validate our approach, we used a dataset used, consist of 15 separate datasets obtained from different PMUs, relay snort alarms and logs. Three machine learning models: Random Forest, Logistic Regression, and K-Nearest Neighbour were built and evaluated using various performance metrics. The findings indicate that the Random Forest model achieves the highest performance with an accuracy of 90.56\% in detecting power system disturbances and has the potential in assisting operators in decision-making processes.
\end{abstract}



\maketitle

\section{Introduction} \label{sec:intro}
Although Cyber-physical system has many advantages in areas such as power distribution grids and wastewater treatment plants, it also has some disadvantages and threats. A smart grid is an electrical grid equipped with automation, communication, and information technology systems that can monitor power flows from points of generation to points of consumption \cite{Paper1}. If these systems fail, it can result in massive damage or loss to people as well as the shutdown of all infrastructure. 

Nowadays, most businesses have regulations and policies in place to ensure their security. Phasor Measurement Units (PMUs) have been used to increase system performance as power systems become increasingly complex in their architecture \cite{paper2}. It provides information that can help to make quick decisions. Hackers, on the other hand, can create a trigger that will cause the system to fail and cause significant damage to smart grids. Machine learning techniques can be used to find pattern recognition, learning abilities, and rapid identification of potential security boundaries \cite{paper2}. This paper proposes a machine learning approach for detecting system behaviors by learning from historical data and relevant information. Mainly we present a machine learning-based attack detection model for power systems that can be taught using data and logs collected by PMUs.

To accomplish this, the dataset was preprocessed, for model selection, 10-fold cross-validation was used to build a random forest, logistic regression, and k-Nearest neighbor models, and the results were compared using four performance metrics: f1 macro, recall, accuracy, and precision scores. Furthermore, feature selection was performed, and the results were compared to models without feature selection; the best model found was Random Forest, and finally, optimization of the best model was performed. 

The structure of this paper is as follows: Section \ref{sec:literature} provides an overview of related research in the field. In Section \ref{sec:approach}, we detail our proposed approach by highlighting the conducted data processing, model building, testing various machine learning methods, and experimental results as well as discussing the findings. Lastly, Section \ref{sec:conclusion} offers concluding remarks.

\section{Literature review} \label{sec:literature}
Smart grids, are vulnerable to cyber-attacks due to their reliance on automation, communication, and information technology systems \cite{Paper1}. Hackers target these systems to disrupt the power supply, cause damage, or gain unauthorized access to critical infrastructure. As highlighted \cite{paper6}, the consequences of successful attacks on power systems can be severe, leading to widespread power outages, financial losses, and even endangering public safety. Therefore, there is an urgent need for effective detection and mitigation strategies to protect power systems from cyber threats.

Machine learning techniques have emerged as promising approaches for enhancing the security of power systems. These techniques offer the ability to analyze large volumes of data, detect patterns, and identify anomalies indicative of potential attacks \cite{paper4}. Phasor Measurement Units (PMUs) play a crucial role in this context, as they provide real-time data on power system dynamics, enabling the development of accurate machine learning models \cite{paper4}. By leveraging historical data and logs collected by PMUs, these models can learn system behaviors and detect deviations that may indicate cyber-attacks.

Several intrusion detection systems(IDS) approaches have been proposed for smart grid security, including anomaly-based detection techniques, communication traffic analysis, and leveraging power system theories \cite{paper3} \cite{paper4}\cite{paper5} \cite{paper6} \cite{paper7}. However, these approaches have limitations in terms of detecting different types of attacks, scalability, and capturing invalid changes in the physical system.

In this study, our goal is to utilize machine learning to detect cyber-attacks and accurately classify different types of power system disturbances. We hypothesize that machine learning algorithms can effectively detect disturbances and classify potential security threats in power systems. By addressing the limitations of existing approaches and harnessing the power of machine learning, we aim to enhance the security and resilience of power systems against cyber-attacks.

\section{Proposed approach} \label{sec:approach}
\subsection{Dataset}

\begin{figure}[h!]
    \centering
    \includegraphics[width=.8\linewidth]{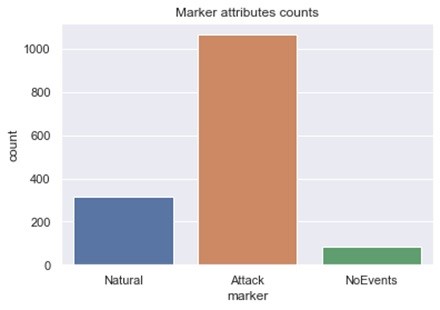}
    \caption{Graph shows the number of each category}
    \label{fig:Marker}
\end{figure}

The dataset downloaded was about power system disturbance. It was made up of 15 separate datasets that were collected and recorded by PMUs 1–4, relay snorts alarms, and logs. Each has 129 columns, and the target attribute was having three classes such as \textit{No even}t, \textit{Natural}, and \textit{Attack} as shown in Figure \ref{fig:Marker}.

\begin{figure}
    \centering
    \includegraphics[width=1\linewidth]{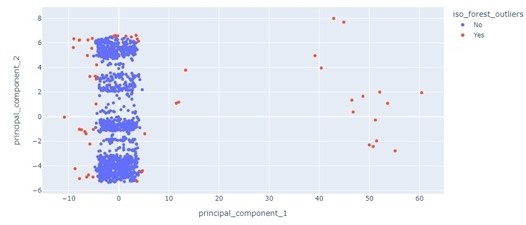}
    \caption{PCA for normal and outliers in the dataset}
    \label{fig:pca}
\end{figure}

All columns were numerical except target attributes which were categorical. Moreover, the total number of observations of all datasets was 73037. After combining all datasets 2 \% from all datasets was collected for this experiment. In the dataset, there were no duplicates or missing values found. However, infinity values were found, and the outlier was detected by using Isolation Forest and Principal component analysis was used to visualize the detected outliers Figure \ref{fig:pca}. 

\subsection{Data preprocessing and preparation}
To prepare the dataset for analysis, several preprocessing steps were performed. Firstly, any infinity values present in the dataset were eliminated. Additionally, outliers were identified using the Isolation Forest algorithm and subsequently removed. To handle non-numerical values, a label encoder was applied to convert them into numerical representations. Moreover, as observed in Figure \ref{fig:Marker}, the dataset exhibited class imbalance. To address this issue, the Synthetic Minority Oversampling Technique (SMOTE) was employed to augment the samples in the minority class. Lastly, to ensure uniformity in the dataset, standardization was carried out by scaling all the features using standard scalers.

\subsection{Exploratory data analysis and data visualization}

\begin{figure}
    \centering
    \includegraphics[width=1\linewidth]{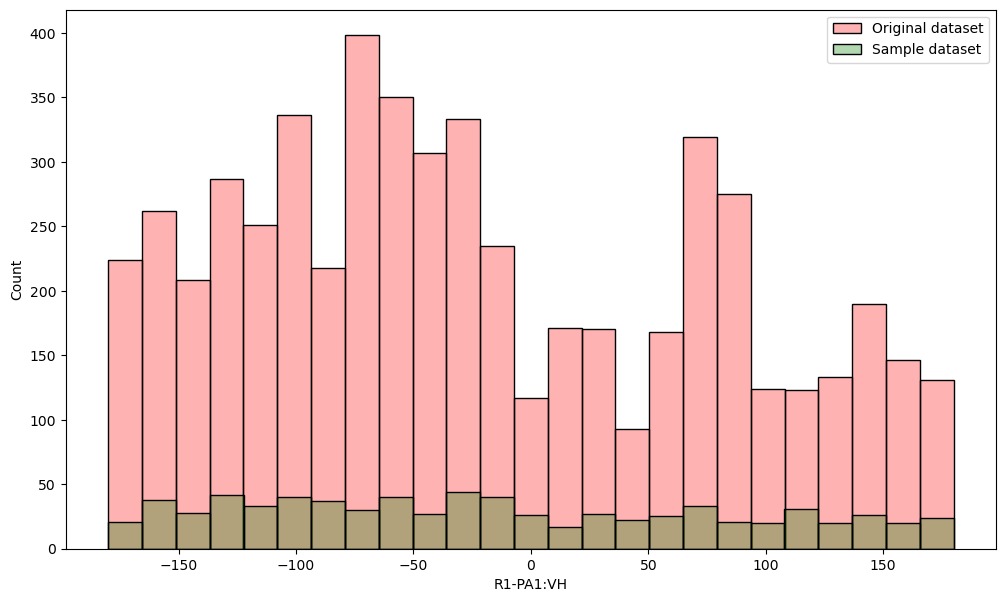}
    \caption{Distribution of the Sampled and original dataset}
    \label{fig:Histogram}
\end{figure}

To explore the data and understand the pattern among features. The distribution of each feature was examined, and an example was presented in Figure  \ref{fig:Histogram} using a histogram. The distribution of the R1-PA1:VH feature closely resembled that of the original dataset, indicating that this particular sample serves as a representative example of the overall dataset.

\begin{figure}
    \centering
    \includegraphics[width=1\linewidth]{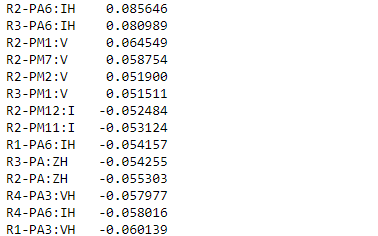}
    \caption{The best 14 features which are very correlated to the Marker}
    \label{fig:tab3}
\end{figure}

Furthermore, correlation analysis was performed to assess the relationships between the features and the target variable. The results were presented in Figure \ref{fig:tab3}, showcasing the most correlated variables. It was found that the top 14 features exhibited strong correlations with the target variable. This suggests that these features hold valuable information and have a significant impact on predicting the target variable. The correlation analysis aids in selecting the most relevant features for subsequent modeling and analysis, ensuring that the chosen variables capture important patterns and relationships within the dataset.

\subsection{Model creation and evaluation}
Three machine learning models, namely Random Forest, Logistic Regression, and K-Nearest Neighbor, were constructed for analysis. To evaluate the performance of each model, 10-fold cross-validation was applied, ensuring robustness and reliable results. Various metrics were used to assess the models, including F1 macro, Precision macro, Recall macro, and Accuracy. Since the dataset underwent resampling to address the class imbalance, these metrics were particularly relevant in evaluating the models' performance on the balanced dataset.

\begin{figure}
    \centering
    \includegraphics[width=1\linewidth]{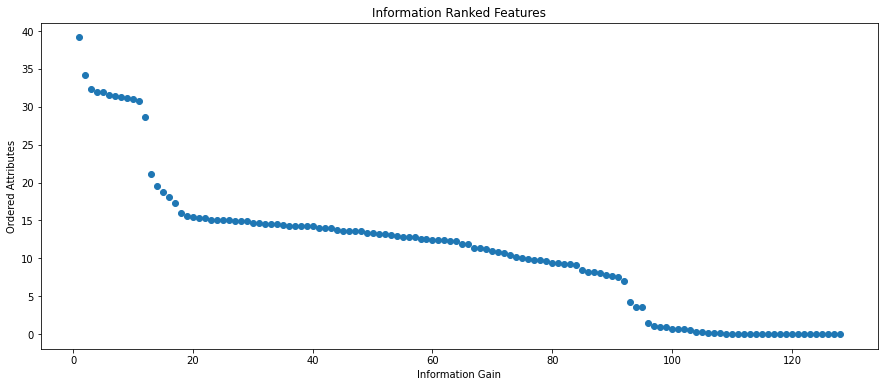}
    \caption{Information gain for each feature}
    \label{fig:mutual_information}
\end{figure}

To determine the impact of feature selection on model performance, the comparison among models was conducted both on the full set of features and after feature selection. The feature selection method employed was mutual information, which measures the dependency of features on the target value. This approach assists in identifying the most informative and relevant features for accurate predictions.  Figure \ref{fig:mutual_information} presents the results of this analysis.

Based on the comparison, the Random Forest model emerged as the best-performing model. Subsequently, hyperparameter tuning was carried out to optimize the selected features. The parameters adjusted during hyperparameter tuning included the number of trees, maximum depth, and criterion selection. By fine-tuning these parameters, the Random Forest model can be optimized to achieve the best possible performance and accuracy for the specific task at hand.

\subsection{Experiments results}

In general, certain features within the dataset were found to exhibit a high correlation with each other, as illustrated in Figure \ref{fig:tab3}. Notably, features such as 'R3-PM9:V', 'R2-PM9:V', 'R4-PM1:V', and 'R3-PM8:V' displayed a strong correlation. However, when considering the correlation between these features and the target variable, the relationship was relatively weaker.

\begin{figure}
    \centering
\includegraphics[width=1\linewidth]{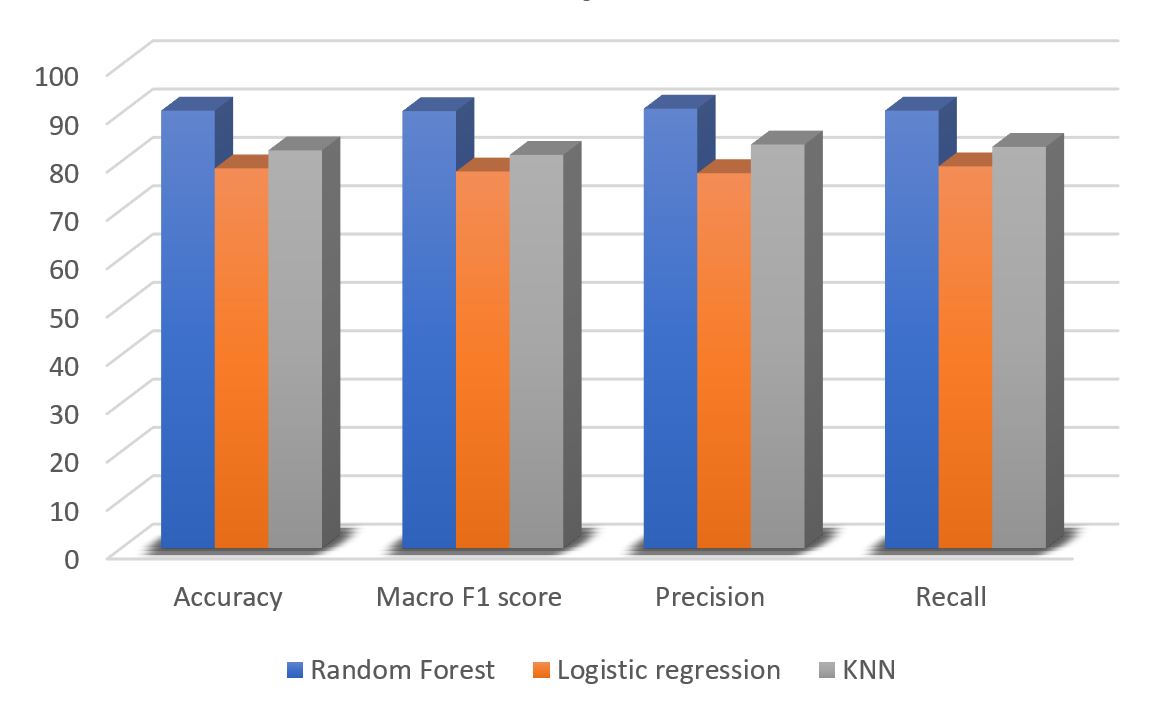}
    \caption{The models' performance without feature selection}
    \label{fig:PerformanceMetrics}
\end{figure}

Additionally, a comparison was conducted among the K-Nearest Neighbor (KNN), Random Forest, and Logistic Regression models. The results demonstrated that the Random Forest model performed the best, achieving an F1 macro score of 90.46\%, an accuracy of 90.56\%, a precision macro score of 90.97\%, and a recall macro score of 90.57\%. Figure \ref{fig:PerformanceMetrics} provides a visual representation of these findings. The second-best performing model was the KNN model, although the specific metrics associated with its performance were not mentioned in the provided context.

Furthermore, the Mutual Information technique was utilized to select the best features from the dataset. Figure \ref{fig:mutual_information} illustrates the scores assigned to each feature based on their relevance. From this analysis, the top 40 features with the highest scores were selected for further modeling.

\begin{figure}
    \centering
\includegraphics[width=1\linewidth]{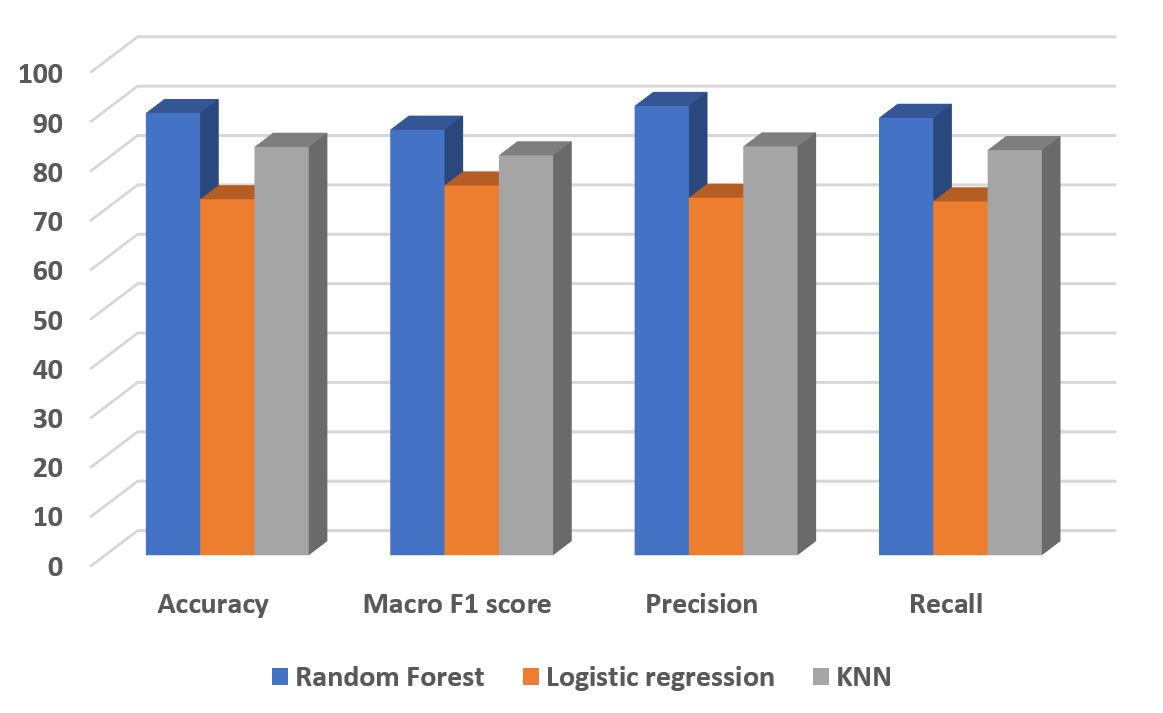}
    \caption{Comparison of the performance of the models with FS}
    \label{fig:performance2}
\end{figure}

Using these selected features, the same machine learning algorithms were constructed and compared once again. The performance of each model was evaluated using metrics such as F1 score, precision, recall, and accuracy, Figure\ref{fig:performance2}. Notably, the Random Forest (RF) model demonstrated strong performance, achieving a Macro F1 score of 86.16\%.

Surprisingly, when comparing the model built with feature selection to the one without, it was found that the model utilizing all features performed better. This unexpected result could be attributed to the potential overfitting of the data since we only used a subset of features.

Additionally, it was observed that the Logistic Regression model did not perform well in this analysis, indicating that it may not be suitable for capturing the complexities present in the dataset or may require further refinement in terms of hyperparameter tuning or feature engineering.
Moreover, the benchmark model found is Random Forest Figure\ref{fig:performance2}, then it was used for Hyperparameter tuning and the accuracy score was improved from 89.54\% to 90.08\%. As a result, it can be concluded that model parameters have to be optimized based on the usage scenario. The model is more sensitive to data collected in the power system and can better distinguish the situations corresponding to the data because of optimization.

\subsection{Discussion}
Previous studies have recommended the application of preprocessing techniques to improve the performance of classifiers, such as balancing the dataset \cite{Paper1}. These findings align with the results obtained in the current study, which also demonstrate that Random Forests exhibit strong precision performance \cite{Paper1}. Furthermore, when comparing different algorithms, the tree-based algorithm Random Forest outperforms KNN and Logistic Regression.

According to Junejo and Goh \cite{paper2}, the success of the Random Forest algorithm can be attributed to the fact that the Programmable Logic Controller (PLC) used in power systems is programmed using relational ladder logic. Ladder logic is a rule-based language that executes rules in sequential order, resembling a control logic system. The tree-based algorithms, including Random Forest, attempt to relearn this control logic or understand the normal behavior of the system. This compatibility between the underlying logic of the power system and the tree-based algorithms could explain the superior performance of Random Forest in this context \cite{paper2}.

\section{Conclusion} \label{sec:conclusion}

This report utilizes Random Forest, KNN, and Logistic Regression machine learning algorithms to detect power system disturbance. All the approaches used for evaluating models showed that random forest remained the best algorithm among others; therefore, it is recommended to be used for classifying the scenarios related to detecting cyberattacks and controlling system operations.  However, an increased amount of data may increase accuracy and time complexity. Moreover, as a recommendation, deep learning and big data can be integrated for future work.

\bibliographystyle{ACM-Reference-Format}
\bibliography{references}
\end{document}